\pdfoutput=1

\documentclass[11pt]{article}

\usepackage[final]{acl}

\usepackage{times}
\usepackage{latexsym}

\usepackage[T1]{fontenc}

\usepackage[utf8]{inputenc}

\usepackage{microtype}

\usepackage{inconsolata}

\usepackage{graphicx}
\usepackage{inconsolata}
\usepackage{inconsolata}
\usepackage{graphicx}
\usepackage{amsmath}
\usepackage{amssymb}
\usepackage{booktabs}
\usepackage{multirow}
\usepackage{booktabs}
\usepackage{pgfplots}
\usepackage{textcomp}
\usepackage{multirow}
\usepackage{mathtools}
\usepackage{amsthm}
\usepackage{csquotes}
\usepackage{subfigure}
\usepackage{xcolor}
\usepackage{enumitem}
\usepackage{pifont}
\usepackage{CJKutf8}
\usepackage{caption}
\usepackage{algorithm}
\usepackage{algpseudocode}
\usepackage{algorithmicx}
\usepackage{enumitem}

\title{DecEx-RAG: Boosting Agentic Retrieval-Augmented Generation with Decision and Execution Optimization via Process Supervision}

\author{
Yongqi Leng\textsuperscript{\rm{1}},
Yikun Lei\textsuperscript{\rm{2}},
Xikai Liu\textsuperscript{\rm{2}},
Meizhi Zhong\textsuperscript{\rm{2}}, 
Bojian Xiong\textsuperscript{\rm{1}}, \\
\textbf{Yurong Zhang}\textsuperscript{\rm{2}},
\textbf{Yan Gao}\textsuperscript{\rm{2}}, 
\textbf{Yi Wu}\textsuperscript{\rm{2}}, 
\textbf{Yao Hu}\textsuperscript{\rm{2}}, 
\textbf{Deyi Xiong}\textsuperscript{\rm{1}}\Thanks{~Corresponding author.} \\
\textsuperscript{1}TJUNLP Lab, College of Intelligence and Computing, Tianjin University, Tianjin, China \\
\textsuperscript{2}Xiaohongshu Inc. \\
\texttt{\{lengyq,xbj1355,dyxiong\}@tju.edu.cn}\\
\texttt{\{leiyikun,zhongmeizhi,liuxikai,zhangyurong\}@xiaohongshu.com}\\
\\
}

\begin{document}
\maketitle
\begin{abstract}
Agentic Retrieval-Augmented Generation (Agentic RAG) enhances the processing capability for complex tasks through dynamic retrieval and adaptive workflows. Recent advances (e.g., Search-R1) have shown that outcome-supervised reinforcement learning demonstrate strong performance. However, this approach still suffers from inefficient exploration, sparse reward signals, and ambiguous global reward feedback.
To address these challenges, we propose DecEx-RAG, which models RAG as a Markov Decision Process (MDP) incorporating decision-making and execution, while introducing an efficient pruning strategy to optimize data expansion. Through comprehensive process-level policy optimization, DecEx-RAG significantly enhances the autonomous task decomposition, dynamic retrieval, and high-quality answer generation capabilities of large language models (LLMs). Experiments show that DecEx-RAG achieves an average absolute performance improvement of $6.2\%$ across six datasets, significantly outperforming existing baselines. Moreover, the pruning strategy improves data construction efficiency by nearly $6 \times$, providing an efficient solution for process-supervised RAG training. The code is available at \url{https://github.com/sdsxdxl/DecEx-RAG}.
\end{abstract}

\section{Introduction}
\label{sec:Introduction}

State-of-the-art LLMs have demonstrated remarkable problem-solving capabilities \citep{DBLP:journals/corr/abs-2412-16720, DBLP:journals/corr/abs-2412-17686, DBLP:journals/corr/abs-2501-12948}. However, due to the inherent limitations of static training data, LLMs still exhibit significant bottlenecks when addressing dynamic and real-time problems \citep{DBLP:journals/csur/JiLFYSXIBMF23, DBLP:journals/tois/HuangYMZFWCPFQL25}. Retrieval-Augmented Generation (RAG) emerges as a promising solution by incorporating external knowledge bases or search engines \citep{DBLP:journals/corr/abs-2312-10997}.

Recent advances employ outcome-supervised reinforcement learning (RL) to integrate dynamic retrieval into reasoning processes, enabling LLMs to autonomously invoke search tools and achieve remarkable results \citep{DBLP:journals/corr/abs-2503-09516, DBLP:journals/corr/abs-2503-05592}. However, this approach still suffers from inherent limitations. First, the efficiency of exploration is low as LLMs must generate a complete reasoning chain before receiving outcome-based reward feedback. Second, reward signals are sparse, which results in the need for more data and training steps to achieve convergence \citep{DBLP:journals/corr/abs-2505-14069}. Third, global rewards struggle to reflect local performance at step levels, posing challenges for fine-grained optimization. 

Inspired by how humans leverage search tools to solve complex problems, we propose DecEx-RAG, a process-supervised framework for Agentic RAG. The framework models RAG as a Markov Decision Process (MDP) consisting of two stages: Decision-Making and Execution.  The Decision-Making stage involves termination and retrieval decisions, the former aims to avoid redundant iterations or premature stopping, while the latter determines whether to rely on the LLMs' internal knowledge or retrieval tools for sub-question resolution.  The Execution stage focuses on the quality of decision execution. For example, a high-quality sub-question is crucial for the success of subsequent reasoning. By structurally decomposing the Decision-Making and Execution stages, our framework enables fine-grained process-level supervision, while taking into account both execution quality and decision efficiency.

\begin{figure*}[t]
\centering
\includegraphics[width=0.95\linewidth]{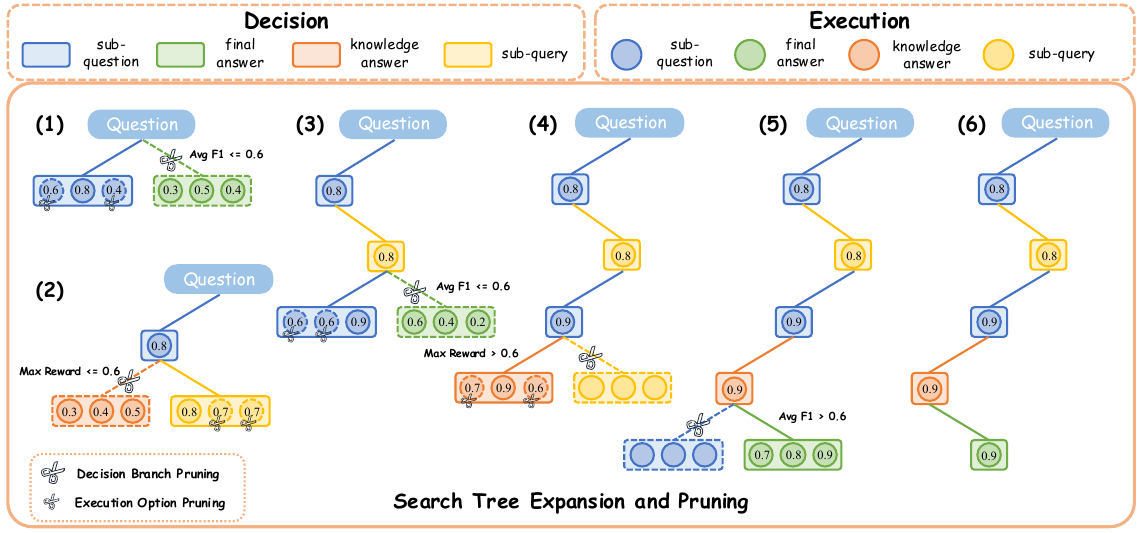}
\caption{Illustration of the framework for DecEx-RAG, which demonstrates the process of search tree expansion and pruning.}
\label{fig1}
\end{figure*}

Within the DecEx-RAG framework, the absence of supervision signals for intermediate reasoning steps often necessitates extensive exploration during search tree expansion, resulting in exponential growth of time complexity with the depth of layers. To address this, we propose an efficient pruning strategy. At each layer of the search tree, multiple rollout simulations are conducted for different executions of decisions, using aggregated rewards as feedback signals to dynamically prune redundant branches. Specifically, the iteration terminates if the termination decision determines that existing iterative information suffices to answer the question; the retrieval operation is skipped if the retrieval decision determines that the model's intrinsic knowledge can effectively derive the final answer.

To validate the effectiveness of the DecEx-RAG framework and the proposed pruning strategy, we conduct experiments on six open-domain QA datasets. Experiment results show that DecEx-RAG significantly outperforms existing methods, achieving an average improvement of $6.3\%$ over the outcome-supervised strong baseline Search-R1 \citep{DBLP:journals/corr/abs-2503-09516}. Further analysis reveals that the proposed pruning strategy enhances data construction efficiency by nearly $6×$ compared to the expansion method without pruning while maintaining equivalent data quality, demonstrating its effectiveness and practicality.

In summary, our main contributions are as follows:
\begin{itemize}[itemsep=1pt, topsep=1pt]
\item We propose DecEx-RAG, a novel framework that models RAG as a Markov Decision Process, providing a more comprehensive and unified perspective for Agentic RAG systems.
\item We propose a pruning strategy to optimize the data expansion process, significantly enhancing the efficiency of process supervision data construction without sacrificing data quality.
\item Experimental results on six datasets demonstrate that, with the same scale of training data, DecEx-RAG significantly outperforms existing baselines in performance.
\end{itemize}

\section{DecEx-RAG}
\label{sec:DecEx-RAG}
Figure \ref{fig1} illustrates our proposed DecEx-RAG framework.  This framework models RAG as a Markov Decision Process incorporating decision-making and execution, expanding a search tree for each question to obtain multi-step process-supervised data. During expansion, intermediate process rewards are acquired through multiple rollouts, while redundant branches are pruned to enhance efficiency. Upon completion of the expansion, the optimal reasoning chain (from root node to leaf node) is used for Supervised Fine-tuning (SFT), while all branched nodes along the path serve as preference data for Direct Preference Optimization (DPO).

\subsection{RAG as a Markov Decision Process}
\label{subsec: RAG as a Markov Decision Process}
We formalize RAG as a Markov Decision Process, represented by the tuple $(\mathcal{S}, \mathcal{A}, P, R)$, where $\mathcal{S}$ denote the state sets, $\mathcal{A}$ denote the action sets, $P$ describes the state transition dynamics, and $R$ specifies the reward function. 
\paragraph{States.}
At each time step $t$, the state $s_t\in \mathcal{S}$ represents a partial solution to the question. Formally, the state $s_{t}$ can be formulated as: $s_t=[Q, (q_1, r_1), \cdots, (q_t, r_t)]$, where $Q$ is the question, $q_i$ denotes the $i$-th sub-question, and $r_i$ represents the answer to $q_i$, or the sub-query and retrieved relevant documents.

\paragraph{Actions.}
In state $s_{t}$, the model selects an action $a_t=(\sigma_t, \delta_t) \in \mathcal A$, which consists of two decisions:
\begin{itemize}[leftmargin=*]
\item Termination decision $\sigma_t$: Determines whether to continue iteration. If continuing, the next sub-question $q_{t+1}$ is generated. If terminating, the final answer $o$ is yielded.
\item Retrieval decision $\delta_t$ (when $\sigma_t$ chooses to continue): For the sub-question $q_{t+1}$, $\delta_t$ determines how to answer it. If choosing to answer using the model's own knowledge, DecEx-RAG generates the answer $w_{t+1}$. If invoking the retriever, DecEx-RAG generates the sub-query $e_{t+1}$ and obtains relevant documents $d_{t+1}$.
\end{itemize}

\paragraph{State Transition.}
When the action $a_{t}=(\sigma_{t},\delta_{t})$ is executed in state $s_t$, the state is updated to $s_{t+1}$. Specifically, if the termination decision $\sigma_{t}$ chooses to terminate, the final answer $o$ will be generated to end the iteration, and the termination state is $s_{t+1}=[Q,(q_1,r_1),\cdots,(q_t,r_t),o]$. Otherwise, DecEx-RAG generates the sub-question $q_{t+1}$ and continues to execute the retrieval decision $\delta_t$.

For the retrieval decision $\delta_t$, if choosing to answer based on model's own knowledge, then $r_{t+1}=w_{t+1}$. If choosing to retrieve, then $r_{t+1}=[e_{t+1}, d_{t+1}]$. Finally, the state is updated to $s_{t+1}=[Q, (q_1, r_1), \cdots, (q_t, r_t), (q_{t+1}, r_{t+1})]$.

\paragraph{Rewards.}
The reward function $R(s_t, a_t)$ specifies the expected reward obtained after taking action $a_t$ in state $s_t$. Under our framework, we perform several rollouts for the state-action pair $(s_t, a_t)$, and use the correctness of multiple rollout results as the reward:
\begin{equation}
R(s_t, a_t)=\frac{1}{n}\sum_{i=1}^n v(\text{rollout}_i)
\label{equ:1}
\end{equation}
where $\text{rollout}_i$ is the $i$-th simulated completion of $(s_t, a_t)$, $v(\text{rollout}_i) \in [0,1]$ denotes the correctness score (e.g., F1 score) of the final answer.

Our framework proposes two key innovations. First, we explicitly distinguish between sub-question and sub-query. The significance of this distinction lies in the fact that different retrievers may return drastically different documents for sub-queries with similar semantics but differing details, so sub-query is processed independently for optimization. Second, we decouple decision-making and execution in the MDP framework. Traditional MDP frameworks often treat decision-making and execution as a whole without clear distinction. In DecEx-RAG, decision-making focuses on method selection, while execution focuses on content quality. By separating decision-making from execution, we aim to leverage decision data to improve system efficiency and execution data to optimize content generation quality.

\subsection{Search Tree Expansion with Pruning}
\label{subsec: Search Tree Expansion with Pruning}
With the Markov Decision Process defined in Section \ref{subsec: RAG as a Markov Decision Process}, DecEx-RAG solves each question by constructing a search tree. In state $s_t$, the model first expands the termination decision $\sigma_t$, using a non-zero temperature parameter to sample termination decision multiple times. If more than $50\%$ of the sampling results favor terminating the iteration, the solving process ends. Otherwise, the model generates multiple candidate sub-questions. After deduplication, we perform multiple rollout simulations for each sub-question, take the average score as the intermediate reward, and finally select the branch with the highest reward to expand the next state.

After completing the termination decision expansion, the model proceeds to expand the retrieval decision $\delta_t$. At this stage, the model first generates multiple candidate answers to the sub-question based on internal knowledge, similarly deduplicating them and calculating rewards through multiple rollouts.  If the highest reward exceeds a preset threshold, it indicates that the final answer can be correctly derived from the current state. In this case, the retrieval branch is skipped to reduce computational overhead, and the highest-reward answer is directly selected as the next expanded state. Otherwise, the model generates multiple sub-queries, repeats the aforementioned process, and selects the sub-query with the highest reward as the next expanded state.

\begin{table*}[t]
\centering
\resizebox{0.9\textwidth}{!}{
\begin{tabular}{l|cc|cc|cc|cc|cc|cc|cc}
   \toprule
   \multirow{2.5}{*}{\textbf{Models}} & \multicolumn{2}{c|}{\textbf{HotpotQA}} &  \multicolumn{2}{c|}{\textbf{2Wiki}} &  \multicolumn{2}{c|}{\textbf{Bamboogle}} &   \multicolumn{2}{c|}{\textbf{PopQA}} &  \multicolumn{2}{c|}{\textbf{NQ}} &  \multicolumn{2}{c|}{\textbf{AmbigQA}} &  \multicolumn{2}{c}{\textbf{Avg}}\\
   \cmidrule(r){2-3} \cmidrule(r){4-5} \cmidrule(r){6-7} \cmidrule(r){8-9} \cmidrule(r){10-11} \cmidrule(r){12-13} \cmidrule(r){14-15} 
   ~ & \textbf{EM} & \textbf{F1} & \textbf{EM} & \textbf{F1} & 
   \textbf{EM} & \textbf{F1} & \textbf{EM} & \textbf{F1} & 
   \textbf{EM} & \textbf{F1} & \textbf{EM} & \textbf{F1} & 
   \textbf{EM} & \textbf{F1} \\
   \midrule
   \multicolumn{15}{l}{\emph{Prompt-Based Methods}} \\
   Direct Inference & 15.7 & 25.3 & 24.2 & 30.7 & 7.2 & 16.2 & 14.3 & 18.6 & 10.5 & 22.4 & 11.6 & 22.4 & 13.9 & 22.6 \\
   CoT & 19.8 & 28.0 & 27.8 & 33.9 & 10.4 & 20.4 & 16.0 & 17.6 & 22.0 & 30.9 & 27.6 & 38.1 & 20.6 & 28.2 \\ 
   Standard RAG & 29.1 & 40.0 & 23.4 & 30.4 & 12.8 & 22.8 & 40.8 & 48.1 & 33.6 & 43.3 & 44.3 & 56.7 & 30.7 & 40.2 \\
   Iter-RetGen & 31.0 & 42.2 & 24.3 & 31.1 & 14.4 & 23.9 & 42.5 & 49.3 & 34.5 & 44.2 & 47.0 & 58.8 & 32.3 & 41.6 \\ 
   IR-COT & 20.5 & 30.9 & 9.1 & 17.2 & 16.0 & 27.9 & 32.4 & 39.9 & 19.3 & 35.5 & 24.5 & 40.6 & 20.3 & 32.0 \\ 
   FLARE & 22.3 & 31.8 & 25.8 & 32.3 & 15.2 & 24.6 & 36.8 & 44.9 & 28.9 & 43.2 & 40.6 & 50.1 & 28.3 & 37.8 \\ 
   Search-o1 & 29.1 & 39.1 & 29.2 & 35.3 & \underline{30.4} & 39.9 & \underline{47.0} & \underline{50.0} & 30.3 & 40.7 & 42.5 & 53.4 & 34.8 & 43.1 \\ 
   \midrule
   \multicolumn{15}{l}{\emph{Reinforcement Learning}} \\
   Search-R1 (PPO) & \underline{32.6}  & \underline{43.2}  & 34.7  & 41.8  & \underline{30.4}  & 43.2  & 41.3  & 46.4  & \textbf{36.0}  & \underline{45.0}  & \underline{49.2}  & \textbf{60.4}  & \underline{37.4}  & \underline{46.7}  \\
   IKEA (GRPO) & 31.1  & 41.7  & 32.7  & 39.3  & \underline{30.4}  & \underline{45.3}  & 38.7  & 42.7  & 30.7  & 42.8  & 47.0  & 57.9  & 35.1  & 45.0  \\
   ReasonRAG & 27.5  & 36.4  & 30.3  & 35.8  & 22.4  & 29.1  & 41.1  & 44.4  & 28.1  & 38.9  & 39.7  & 51.9  & 31.5  & 39.4  \\
   DeepRAG\textsuperscript{\dag} & 32.1  & 41.1  & \underline{40.4}  & \underline{44.9}  & - & - & 40.6  & 43.2  & - & - & - & - & -  & -  \\
   \textbf{DecEx-RAG} & \textbf{37.7}  & \textbf{49.3}  & \textbf{50.0}  & \textbf{55.9}  & \textbf{37.6}  & \textbf{49.3}  & \textbf{51.3}  & \textbf{53.2}  & \textbf{36.0}  & \textbf{47.2}  & \textbf{49.5}  & \underline{59.5}  & \textbf{43.7}  & \textbf{52.4} \\
   \bottomrule
\end{tabular}}
\caption{The overall experimental results of DecEx-RAG and other baselines on six datasets. The best/second best scores in each dataset are \textbf{bolded}/\underline{underlined}. \textsuperscript{\dag}DeepRAG EM/F1 scores are extracted from the DeepRAG paper \citep{DBLP:journals/corr/abs-2502-01142}.}
\label{tab_1}
\end{table*}

The search tree expansion process is controlled by the maximum iteration limit $T_{\max}$. Each layer executes according to the above strategy until either reaching the iteration limit or being actively terminated by the model. Although each decision execution requires multiple rollouts to obtain intermediate rewards, these data not only serve to construct preference training data but also provide the basis for decision pruning. Through pruning, each layer of the search tree retains only the optimal decision branch, significantly enhancing the expansion efficiency of the search tree. All the prompt instructions used for the search tree expansion can be found in Appendix \ref{appendix-sec:Prompt Instructions}.

\subsection{Training}
DecEx-RAG focuses on optimizing four key components: final answer generation, sub-question decomposition, sub-question answering based on self-knowledge, and sub-query generation and retrieval, covering all branches of the framework's two decisions. Based on the search tree constructed in Section \ref{subsec: Search Tree Expansion with Pruning}, we perform two-stage training.

\paragraph{Supervised Fine-tuning (SFT)} We extract the complete reasoning path from the root node to the leaf node in the search tree for Supervised Fine-Tuning. For each question Q, we construct multiple input-output pairs, where the input consists of $\{Q,(q_1, r_1), ...,(q_{x_i}, e_{x_i}, d_{x_i})\}$, and the output consists of ${(q_{x_{i}+1}, r_{x_{i}+1}), ...,(q_{x_{i+1}}, e_{x_{i+1}})}$, with $x_i$ is the index of the retrieval step. This setup ensures iterative reasoning, enabling LLMs to leverage previous steps until the next retrieval or final answer generation.

\paragraph{Direct Preference Optimization (DPO)} Subsequently, we utilize the decision and execution data from the search tree nodes to construct preference pairs. Although only the optimal branch is retained for each layer of the final search tree, all decision branches are generated before pruning. As a result, we naturally obtain two types of preference data. Based on the mixed preference data, we perform DPO training, and the detailed training objective is shown in Appendix \ref{appendix-sec:DPO Training Objective}.

\section{Experiments}
\label{sec:Experiments}

\subsection{Datasets and Baselines}
We evaluated DecEx-RAG and all baseline models on six public datasets, including three single-hop QA datasets: PopQA \citep{DBLP:conf/acl/MallenAZDKH23}, NQ \citep{DBLP:journals/tacl/KwiatkowskiPRCP19}, and AmbigQA \citep{DBLP:conf/emnlp/MinMHZ20}, and three multi-hop QA datasets: HotpotQA \citep{DBLP:conf/emnlp/Yang0ZBCSM18}, 2WikiMultiHopQA \citep{DBLP:conf/coling/HoNSA20}, and Bamboogle \citep{DBLP:conf/emnlp/PressZMSSL23}. Among these, HotpotQA and 2WikiMultiHopQA are in-domain test sets, while the other four are out-of-domain test sets.
We compared DecEx-RAG against two categories of baselines: prompt-based approaches (e.g., Direct Inference, Standard RAG, Search-o1) and reinforcement learning approaches (e.g., Search-R1, IKEA, DeepRAG). Detailed baseline information can be found in Appendix \ref{appendix-sec:Baselines}. 

\begin{figure*}[t]
  \centering
    \subfigure[]{\label{fig2:a}
       \centering
       \includegraphics[width=0.31 \textwidth]{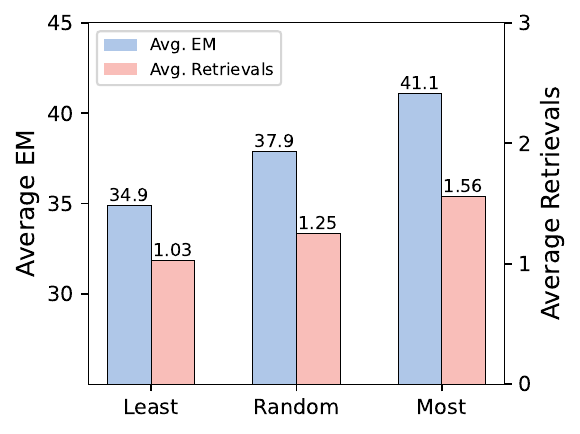}
    }
    \subfigure[]{\label{fig2:b}
        \centering
        \includegraphics[width=0.31 \textwidth]{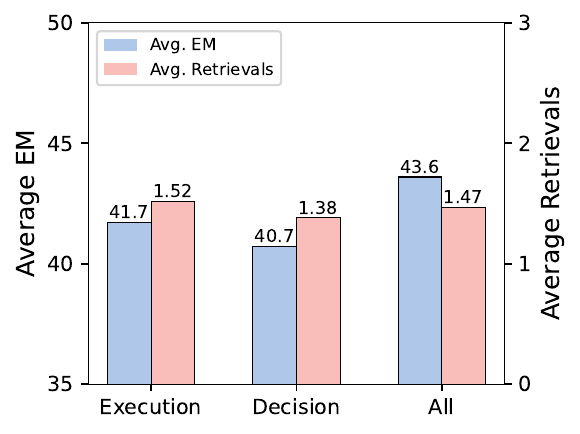}
    }
    \subfigure[]{\label{fig2:c}
        \centering
        \includegraphics[width=0.31 \textwidth]{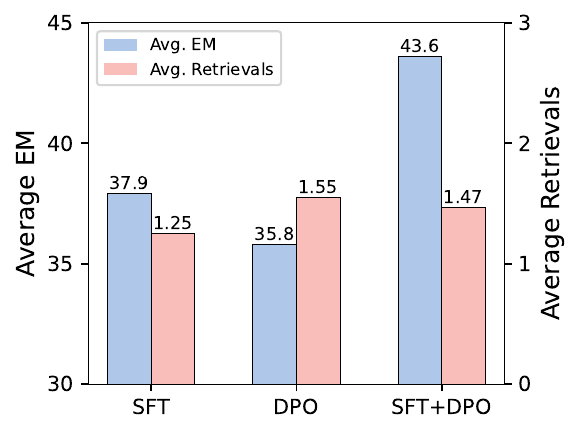}
    }
  \caption{Ablation results for SFT (a), DPO (b) and different training methods (c).
  }
  \label{fig2}
\end{figure*}

\subsection{Implementation Details}
We extracted 2,000 and 1,000 questions from the training subsets of HotpotQA and 2WikiMultiHopQA, respectively, to construct our training dataset. During the expansion of the search tree, for retrieval decisions, we employed Qwen2.5-7B-Instruct \citep{DBLP:journals/corr/abs-2412-15115} as the policy model, which is consistent with the base model in the subsequent training stage. Apart from this decision, we uniformly used Qwen3-30B-A3B \citep{DBLP:journals/corr/abs-2505-09388} as the policy model. For retrieval, we follow \citet{DBLP:journals/corr/abs-2503-09516}, using the 2018 Wikipedia dump \citep{DBLP:conf/emnlp/KarpukhinOMLWEC20} as the knowledge source and E5 \citep{DBLP:journals/corr/abs-2212-03533} as the retriever.

To ensure fair comparison, we reproduced Search-R1 and IKEA with equivalent amounts of training data.  The number of retrieved documents is set to $3$ for all methods requiring retrieval, while $T_{\max}$ is set to $4$ for all multi-round iterative methods.  For evaluation, Exact Match (EM) and F1 scores are adopted as the metrics.

\subsection{Main Results}
As shown in Table \ref{tab_1}, we present the performance of DecEx-RAG and $11$ baseline models on six datasets. Based on the experimental results, we draw the following main conclusions:
\paragraph{Prompt-Based Methods Exhibit Limited Performance.} Experiments show that methods relying solely on internal knowledge (e.g., Direct Inference and CoT) show poor performance, confirming the limitations of LLMs' inherent knowledge. In contrast, RAG-based methods (e.g., Iter-RetGen, IR-COT, FLARE) improve performance, highlighting the necessity of external knowledge integration. Notably, although Search-o1 achieves the best performance among prompt-based methods, this lead is far smaller than the significant advantage demonstrated in its paper.  This is because Search-o1's complex prompting pipeline imposes high requirements on the policy model's instruction-following and reasoning capabilities, and its advantages only manifest fully on large reasoning models. Consequently, it is difficult to extend this advantage equally to smaller models.
\paragraph{Process-Supervised RL Demonstrates Higher Data Efficiency Than Outcome-supervised RL.} While Search-R1 and IKEA improve the LLMs' ability to integrate search during reasoning through outcome-supervised RL, their performance remains suboptimal with only 3K training samples. In contrast, DecEx-RAG demonstrates comprehensive leadership across six datasets, yielding an average performance improvement of $6\%$\textasciitilde$8\%$ over outcome-supervised RL methods under the same training data scales, fully demonstrating the data efficiency advantage of process-supervised RL.
\paragraph{DecEx-RAG Outperforms Other Process-Supervised Methods.} Compared with similar process-supervised methods (DeepRAG and ReasonRAG), DecEx-RAG shows better performance. DeepRAG only focuses on decision optimization, while DecEx-RAG covers both decision and execution, forming a more complete optimization framework. Output analysis of ReasonRAG shows that its insufficient retrieval behavior and excessive reliance on internal knowledge lead to more incorrect answers. Experimental results show that DecEx-RAG not only performs outstandingly on in-domain test sets but also demonstrates excellent cross-domain generalization ability.

\subsection{Ablation Study}
To validate the effectiveness of the DecEx-RAG framework, we conducted a series of ablation experiments. Figure \ref{fig2:a} shows comparative results of different data selection strategies during the SFT stage.   Experiments show that the model trained with the Most Retrieval Cost (Most) strategy exhibits optimal performance, with more frequent retrieval actions.   By examining the model's outputs, we find that the model tends to verify answer accuracy through multiple retrievals, demonstrating stronger deliberative capabilities.  In contrast, the model trained with the Least Retrieval Cost (Least) strategy overly relies on its own knowledge, leading to increased error rates.

\begin{table*}[t]
\centering
\resizebox{0.75\textwidth}{!}{
\begin{tabular}{lccc}
   \toprule
    \textbf{Metrics} & \textbf{Pruning Search} & \textbf{No Pruning Search} & \textbf{Full Node Search} \\
   \midrule
   \multicolumn{4}{c}{\emph{Expansion Efficiency $(k=3, n=4, T_{\max} =4)$}} \\
   \midrule
   Theoretical Expansion Counts & $(4k+3knl)l$ & $\sum_{i=1}^l{i^2(4k+3knl)}$ & $(2k(k+1))^l$  \\
   Average Extension Time (s) & 134.9 & 743.2 & - \\
   \midrule
   \multicolumn{4}{c}{\emph{Data Quality $(k=3, n=4, T_{\max} =4)$}} \\
   \midrule
   SFT (EM/F1) & 32.8 / 40.7 & 33.2 / 41.8 & - \\
   SFT+DPO (EM/F1) & 36.6 / 45.3 & 36.3 / 44.8 & - \\
   \bottomrule
\end{tabular}}
\caption{Comparison of the three expansion methods. $k$ denotes the number of execution branches for each decision, $n$ denotes the number of rollouts and $l$ denotes the depth of layers.}
\label{tab_2}
\end{table*}

Figure \ref{fig2:b} shows the comparison of different preference data compositions during the DPO stage. Experiments reveal that omitting any type of preference data results in performance degradation, highlighting the necessity of simultaneously optimizing decision-making and execution.  Further analysis shows that the model trained with decision data performs slightly worse than the one trained with execution data but requires fewer retrievals. This confirms that execution data primarily optimizes content quality, while decision data optimizes retrieval efficiency.

Figure \ref{fig2:c} shows performance comparisons across different training methods. Experiments reveal that SFT-only training enables models to learn basic retrieval patterns, but insufficient retrieval frequency limits performance ceilings. DPO-only trained models suffer from poor iteration quality and increased ineffective retrievals due to the lack of effective imitation learning. In comparison, the two-stage training (SFT+DPO) achieves a balance between performance and efficiency, establishing reasoning foundations through SFT training while optimizing decision processes via DPO training, ultimately delivering the best overall performance.

\section{Analysis}
\subsection{Data Expansion Efficiency}
To validate the effectiveness of the search tree expansion pruning method (Pruning Search), we compared it with two other approaches: No Pruning Search and Full Node Search. No Pruning Search retains the optimal execution branches for the two decisions, while Full Node Search keeps all execution branches but omits the rollout simulation process.  Table \ref{tab_2} reports all the results.

Theoretical analysis shows that the number of expansions in Pruning Search grows linearly with the depth of layers, significantly outperforming the exponential growth trend of $2^l$ in No Pruning Search and the more drastic growth trend of $(2k)^l$ in Full Node Search, where $k$ represents the number of execution branches per decision, $l$ represents the depth of layers.

In the actual performance test, we evaluated the three methods using $500$ questions under the same hardware configuration.  Experimental results show that Pruning Search is nearly $6$ times faster than No Pruning Search in terms of average expansion time per question.  The expansion efficiency of Full Node Search is extremely low, with the expansion time per question exceeding $1$ hour, making it incomparable.  Therefore, we do not report its subsequent experimental results.

To further validate the data effectiveness, we sampled data generated by Pruning Search and No Pruning Search in the same quantity and used it to train Qwen2.5-7B-Instruct, respectively. Table \ref{tab_2} shows the average performance across six datasets. We observe that whether SFT-only training or SFT+DPO training, the data generated by Pruning Search and No Pruning Search yield nearly identical model performance. 

In summary, the pruning strategy significantly improves expansion efficiency (nearly $6$ times faster) while maintaining data quality comparable to No Pruning Search, fully demonstrating the effectiveness and practical value.

\subsection{Effect of Pruning on Optimal Reasoning Chain}
In general, pruning operations may bias the search toward locally optimal reasoning paths, while discarding potentially superior global reasoning paths. For RAG tasks, the core differences between different reasoning paths for the same question usually lies in the choice made during the decision-making process. Thus, we examined the optimal reasoning chains generated by Pruning Search and No Pruning Search.  

We find that for $85$\% of the samples, the reasoning chains obtained by these two methods have the same number of iterations, and for $87$\% of the samples, the reasoning chains obtained by the two methods have the same number of retrievals. These results directly demonstrate that the proposed pruning strategy can maximally preserve the globally optimal reasoning chain while compressing the search space.

\subsection{Case Study}
We provide a case in Table \ref{tab_3} to intuitively compare the performance differences between DecEx-RAG and Search-R1. 

For the given question, Search-R1 relying on its own knowledge and one retrieval, accurately clarified the nationality information of Ed Wood and Scott Derrickson during the reasoning process. However, its final output result is ``No''.  This phenomenon clearly indicates that Search-R1 has a serious inconsistency between the reasoning process and the conclusion, which is a typical case of reward hacking. In contrast, DecEx-RAG not only generated a logically rigorous and correct reasoning process, but also achieved a high degree of consistency between the reasoning process and the final answer. 

\section{Related Work}
\label{Related_Work}
\paragraph{RL for Retrieval-Augmented Generation}
Recent advances show that reinforcement learning (RL) significantly enhances LLMs' complex reasoning, emerging as a mainstream post-training paradigm \citep{DBLP:journals/corr/abs-2501-12948, DBLP:conf/coling/ZhangX25}. Agentic RAG like Search-R1 \citep{DBLP:journals/corr/abs-2503-09516} use outcome-supervised RL to train LLMs as search agents, improving complex task performance \citep{DBLP:journals/corr/abs-2503-05592, DBLP:journals/corr/abs-2503-19470}.  However, this approach faces inefficient exploration, sparse reward signals, and ambiguous global feedback.  In contrast, Process-supervised RL incorporates fine-grained step-level rewards and has outperformed outcome supervision in various tasks \citep{DBLP:conf/iclr/LightmanKBEBLLS24}. Nevertheless, its integration into RAG training pipelines remains under-explored.

\paragraph{Tree Search for LLMs}
Tree search have been extensively explored in both the training and inference of LLMs, particularly in data synthesis \citep{DBLP:conf/nips/ZhangZHYD024, DBLP:conf/acl/YuJLKSZZWZX25} and inference-time scaling \citep{ DBLP:journals/corr/abs-2408-03314, DBLP:conf/acl/LiPWCSZSWWWWHLW25}. \citet{DBLP:conf/nips/YaoYZS00N23} propose Tree of Thoughts (ToT), which generates multiple branches for substeps during inference to expand reasoning paths. Similarly, several studies incorporate Monte Carlo Tree Search (MCTS) during reasoning to improve the performance of LLMs in mathematical and RAG-related tasks \citep{DBLP:conf/nips/ChenL0024a, DBLP:conf/naacl/JiangCLRWZSZ25, feng2025airrag}. While inference-time scaling can enhances the performance of LLMs, it significantly increases computational complexity, limiting practical applications. Considering this constraint, \citet{DBLP:conf/nips/ZhangDPLGL24} propose fine-tuning LLMs by collecting preference information during tree search processes, effectively transferring resource consumption from inference stage to training stage. However, tree search still relies on extensive search and simulation operations, and its search efficiency and scalability in RAG remain to be optimized.

\section{Conclusion}
In this paper, we have presented DecEx-RAG, a novel framework that models RAG as a Markov Decision Process (MDP) with two distinct stages: Execution and Decision-Making. By decomposing the process into structured steps, DecEx-RAG enables fine-grained process supervision, facilitating more precise optimization of RAG systems. Based on this framework, we have introduced a search tree pruning strategy that incorporates multiple additional rollout processes to provide reward signals for intermediate steps. This allows the framework to dynamically prune unnecessary decision branches, effectively reducing the time complexity of search tree expansion from exponential to linear. Experiments on six datasets have demonstrated that DecEx-RAG consistently outperforms existing baselines. Further analysis has confirmed the efficiency of the pruning strategy, achieving up to a $6×$ improvement in data construction efficiency without sacrificing data quality. This offers a practical solution to the high cost of process-supervised data construction.

\section*{Limitations}
Despite the outstanding performance of DecEx-RAG, several limitations remain. When constructing the process supervision data, we adopt Exact Match (EM) and F1 score as reward feedback for intermediate processes. However, we have found that in some samples where errors occur in intermediate steps, high scores can still be obtained after rollout.  This phenomenon is mainly attributed to the fact that answers in existing RAG datasets are generally too concise and problems are simplistic, leading to EM and F1 score failing to truly reflect the correctness of responses in some scenarios. Currently, we mitigate the inconsistency between intermediate processes and final results by increasing the number of rollouts, but this approach substantially increases reasoning resource consumption.  

In the future, we plan to explore a more scientifically authoritative evaluation metric for RAG tasks, rather than merely relying on increasing the frequency of rollouts, we aim to fundamentally enhance the accuracy and reliability of intermediate process feedback.

\section*{Acknowledgments}
The present research was supported by the Key Research and Development Program of China (Grant No. 2023YFE0116400). We would like to thank the anonymous reviewers for their insightful comments.

\bibliography{custom}

\appendix
\clearpage

\section{Method Details}

\subsection{Prompt Instructions}
\label{appendix-sec:Prompt Instructions}
We use several prompt instructions during the search tree extension process. Table \ref{tab_A.1-1} provides the prompt instruction of execution termination decision, Table \ref{tab_A.1-2} provides the prompt instruction for generating sub-question, Table \ref{tab_A.1-3} presents the prompt instruction for the model to answer questions based on its own knowledge, Table \ref{tab_A.1-4} provides the prompt instruction for generating sub-query, and Table \ref{tab_A.1-5} presents the prompt instruction for rollout.

\subsection{DPO Training Objective}
\label{appendix-sec:DPO Training Objective}
The DPO training objective is formulated as:
\begin{equation}\small
\begin{aligned}
\mathcal L_\theta = - \mathbb{E}_{(x, y_{<t}, y_t^{w}, y_t^{l}) \sim \mathcal{D}} 
\Bigg[\log &\sigma \Big( \beta \log \frac{\pi_{\theta}(y_t^w | x, y_{<t})}{\pi_{\text{ref}}(y_t^w | x, y_{<t})} \\ & - \beta \log \frac{\pi_{\theta}(y_t^l | x, y_{<t})}{\pi_{\text{ref}}(y_t^l | x, y_{<t})}\Big)\Bigg]
\end{aligned}
\end{equation}
where $\pi_{\theta}$ represents the policy being optimized, $\pi_{ref}$ represents the reference policy, $x$ denotes the original question, $y_{<t}$ denotes the previous reasoning steps, $y_{t}^{w}$ and $y_{t}^{l}$ respectively represent the preferred and non-preferred decisions or executions in the subsequent steps, $\sigma$ is the logistic function, and the hyperparameter $\beta$ controls the KL divergence constraint.

\section{Baselines}
\label{appendix-sec:Baselines}
The baselines we compared against are as follows:
\begin{itemize}
\item Direct Inference: We prompt the model to directly answer questions.
\item CoT: We prompt the model to reason step-by-step before generating the answer.
\item Standard RAG \citep{DBLP:conf/nips/LewisPPPKGKLYR020}: We first retrieve relevant documents for the question, then prompt the model to answer with reference to these documents.
\item Iter-RetGen \citep{DBLP:conf/emnlp/ShaoGSHDC23}: This method collaboratively retrieves and generates in an iterative manner, alternating between optimizing retrieval and generation through multiple iterations.
\item IRCoT \citep{DBLP:conf/acl/TrivediBKS23}: This method interweaves retrieval with CoT reasoning steps, using CoT to guide retrieval and leveraging retrieval results to refine CoT.
\item FLARE \citep{DBLP:conf/emnlp/JiangXGSLDYCN23}: This method introduces proactive forward-looking retrieval, determining retrieval timing based on token-level confidence.
\item Search-o1 \citep{DBLP:journals/corr/abs-2501-05366}: This method designs a complex iterative workflow via prompting and introduces a Reason-in-Documents module to optimize reasoning chains.
\item Search-R1 \citep{DBLP:journals/corr/abs-2503-09516}: This method trains the model's dynamic retrieval capabilities via outcome-supervised reinforcement learning.
\item IKEA \citep{DBLP:journals/corr/abs-2505-07596}: Extending Search-R1. This method introduces a retrieval frequency penalty to improve reasoning efficiency.
\item ReasonRAG \citep{DBLP:journals/corr/abs-2505-14069}: This method constructs process-supervised data via Monte Carlo Tree Search (MCTS) for DPO training.
\item DeepRAG \citep{DBLP:journals/corr/abs-2502-01142}: This method constructs process-supervised data via full-node expansion combined with binary tree search for DPO training.
\end{itemize}

\begin{table*}
\centering\small
\begin{tabular}{p{14cm}}
\toprule
\textbf{Question:} Were Scott Derrickson and Ed Wood of the same nationality? \\
\midrule
\textbf{Ground Truth:} yes \\
\midrule
\textbf{DecEx-RAG:} \\
<think> To determine if Scott Derrickson and Ed Wood share the same nationality, I need to identify the nationalities of both individuals. Starting with Scott Derrickson. So, the next sub-question is: What is Scott Derrickson's nationality? I recall that Scott Derrickson's nationality is American. Now, I need to determine Ed Wood's nationality. I am less familiar with him, so I require additional information. I am uncertain about Ed Wood's nationality. He may be an actor or filmmaker, but I need to verify this. </think> \\

<search> Ed Wood nationality </search> \\

<information> \\
Docs 1: "Ed Wood" \\
Grey\'s 1992 oral biography "Nightmare of Ecstasy: The Life and Art of Edward D. Wood Jr.", the public took a greater interest in Wood, leading up to director Tim Burton's biopic of Wood's life, "Ed Wood" (1994), a critically acclaimed film starring Johnny Depp as Wood that earned two Academy Awards. Wood's father, Edward Sr., worked for the U.S. Postal Service as a custodian, and his family relocated numerous times around the United States. Eventually, they settled in Poughkeepsie, New York, where Ed Wood Jr. was born in 1924. According to Wood's second wife, Kathy O'Hara, Wood's mother Lillian would... \\

Docs 2: "Ed Wood" \\
Ed Wood Edward Davis Wood Jr. (October 10, 1924 – December 10, 1978) was an American filmmaker, actor, and author. In the 1950s, Wood made a number of low-budget films in the science fiction, crime and horror genres, intercutting stock footage. In the 1960s and 1970s, he made sexploitation movies and wrote over 80 pulp crime, horror and sex novels. In 1975, he was awarded a Golden Turkey Award as Worst Director of All Time, renewing public interest in his work. Wood's career and camp approach has earned him and his films a cult following. Following the publication of Rudolph... \\

Docs 3: "Alexander Wood (soccer)" \\
Alexander Wood (soccer) Alexander “Alec” Wood (June 12, 1907 – July 20, 1987) Was a former Scottish American soccer defender. Wood began his club career in the United States before moving to England in the early 1930s. He also played all three U.S. games at the 1930 FIFA World Cup. He is a member of the National Soccer Hall of Fame. Wood’s parents moved the family to the United States in 1921 when Wood was fourteen years old. His family settled in Gary, Indiana where he attended Emerson High School and gained his U.S. citizenship a year later. He also... \\
</information>

<think> The information provided confirms that Ed Wood was an American filmmaker, and his birth in the U.S. establishes his nationality as American. Scott Derrickson is also American. Both are of the same nationality. </think> \\
<answer> yes </answer> \\
\midrule
\textbf{Search-R1:} \\
<think> Scott Derrickson is known for directing films such as "The Exorcist" (2016). Ed Wood is famous for his low-budget films in the 1950s, particularly "Plan 9 from Outer Space." Given that Ed Wood was American, I need to confirm Scott Derrickson\'s nationality. </think> \\

<search> Scott Derrickson nationality </search> \\
<information> \\
Docs 1: "Scott Derrickson" \\
directed the film "Doctor Strange", based on the Marvel Comics property and part of the Marvel Cinematic Universe. It was released in November 2016. The film was both a commercial and critical success. In February 2018, it was announced that Derrickson will executive produce the television series "You Bury Me" for Blumhouse Productions. The series is set in Iraq, Syria, and Turkey, and will be a love story in contemporary war-torn Middle East. In December 2018, it was announced that Derrickson will direct the "Doctor Strange" sequel with a planned May 2021 release. Derrickson married a nurse in September 1993... \\

Docs 2: "Scott Tucker (racing driver)" \\
(in the TA2 class) in August 2011 at Road America. Tucker is the founder and CEO of Sentient Technologies, a provider of loan origination software. Tucker grew up in Kansas City, Missouri and went to Rockhurst High School. He studied business administration at Kansas State University. Tucker is married to his wife Kim and has two daughters. On January 5, 2018, Tucker was convicted of 14 criminal counts relating to his payday lending businesses and was sentenced to 16 years and 8 months in prison. Tucker is presently in the custody of the U.S. Marshals Service at an undisclosed location. \\

Docs 3: "Scott Smith (ice hockey)" \\
collaborated regularly with team management of Sean Burke with the men's national team, and Melody Davidson with the women\'s national team to ensure that staff and athletes were prepared to perform on the ice. In 2018, Smith extended the partnership with Bauer Hockey for another 8 years, who have been the official equipment supplier for the national teams since 1996. Smith was born November 27, 1966, in Bathurst, New Brunswick. He currently resides in Calgary, and is married to Karen, with one son named Jackson, who grew up playing hockey. Smith has family in the Bathurst area, including in-laws, and... \\
</information> \\

<think> From the search results, it appears that Scott Derrickson is associated with American films and there is no mention of him being from any other country. Therefore, Scott Derrickson is American. </think> \\
<answer> No </answer> \\
\bottomrule
\end{tabular}
\caption{A case study of DecEx-RAG and Search-R1.}
\label{tab_3}
\end{table*}

\begin{table*}
\centering\small
\begin{tabular}{p{14cm}}
\toprule
\textbf{Prompt Instruction} \\
\midrule
  
You are a Retrieval-Augmented Generation (RAG) model. Your task is to carry out multiple iterative processes, continuously decompose a problem into sub-problems, solve them one by one, and finally answer the original question.  We will provide you with a question and the previous iteration history (each iteration includes the thinking process, sub-question, the model's answers to the sub-questions based on its own knowledge or the relevant documents retrieved from the sub-question). Please judge based on the available information whether there is sufficient evidence to answer the original question reasonably. Please note that you must rely only on the existing information as evidence and cannot introduce any information or knowledge that has not appeared in the iterative history, including your own knowledge.\\
\\

- If you think that based on the existing information, you can reasonably reason out the answer to the original question and there is no need to continue raising the next sub-question, then please first state your reasoning process in <reasoning> and </reasoning>, the reasoning process requires the provision of reasonable evidence, and then provide the final answer in <answer> and </answer>.  For example: <reasoning> your reasoning process </reasoning>. <answer> your answer </answer>.\\

- If you think that it is impossible to reasonably reason out the answer to the original question based on the existing information and you need to continue to raise the next sub-question and iterate further, then please first state your reasoning process in <reasoning> and </reasoning>, the reasoning process needs to explain the reasons for the need for further iteration, and then provide the sub-questions in <question> and </question>. For example: <reasoning> your reasoning process </reasoning>. <question> next sub-question </question>.\\
\\

Matters Need Attention:\\
**1. It should be noted that if you find that the sub-question generated in the previous iteration has not been successfully solved, we encourage you to try to raise this sub-question again and continue the iteration, rather than easily giving a final answer and thus ending the iteration.**\\
**2. Ensure both question and answer are concise, using nouns or short phrases whenever possible.**\\
\\

\#\#\# Question \\
\{question\} \\
\\

\#\#\# Previous Iteration \\
\{iter\_history\} \\
\\

\#\#\# Your Output \\
\\
\bottomrule
\end{tabular}
\caption{The prompt of execution termination decision determines whether to continue the iteration or terminate.}
\label{tab_A.1-1}
\end{table*}

\begin{table*}
\centering\small
\begin{tabular}{p{14cm}}
\toprule
\textbf{Prompt Instruction} \\
\midrule
You are a Retrieval-Augmented Generation (RAG) model. Your task is to break down the original question into several clear sub-questions, which will be solved iteratively to answer the original question. \\
We will provide you with a question.  Please break it down and output the next sub-question that needs to be solved.  Before outputting sub-question, you must provide a clear thinking process or reasons.  Finally, provide the sub-question inside <question> and </question>. \\
For example: Your thinking process. <question> sub-question </question>. \\
\\
\#\#\# Question \\
\{question\} \\
\\

\#\#\# Your Output \\
\\
\bottomrule
\end{tabular}
\caption{The prompt for generating sub-question.}
\label{tab_A.1-2}
\end{table*}

\begin{table*}
\centering\small
\begin{tabular}{p{14cm}}
\toprule
\textbf{Prompt Instruction} \\
\midrule
We will provide you with a question, please answer this question.  Your output should first provide your thinking process, and finally offer the final answer within <answer> and </answer>, ensure your answer is concise, using nouns or short phrases whenever possible. An example is shown below.\\
\\
\#\#\# Sample\\
Question: \\
Which country's capital is Beijing?\\
\\
Your Output:\\
To determine which country's capital is Beijing, I need to recall geographical knowledge about major cities and their respective countries. Beijing is a well-known city in East Asia, and it is the political and cultural center of a significant nation in that region. The country that has Beijing as its capital is China. <answer> China </answer>\\

\\
\#\#\# Question \\
\{question\} \\
\\

\#\#\# Your Output \\
\\
\bottomrule
\end{tabular}
\caption{The prompt for the model to answer questions based on its own knowledge.}
\label{tab_A.1-3}
\end{table*}

\begin{table*}
\centering\small
\begin{tabular}{p{14cm}}
\toprule
\textbf{Prompt Instruction} \\
\midrule
You are a Retrieval-Augmented Generation (RAG) model. We will provide you with a question. Please generate a query for search engines for this question. Generate the query directly without outputting any other content.\\

\\
\#\#\# Question \\
\{question\} \\
\\

\#\#\# Your Output \\
\\
\bottomrule
\end{tabular}
\caption{The prompt for generating sub-query.}
\label{tab_A.1-4}
\end{table*}

\begin{table*}
\centering\small
\begin{tabular}{p{14cm}}
\toprule
\textbf{Prompt Instruction} \\
\midrule
You are a Retrieval-Augmented Generation (RAG) model.  Your task is to carry out multiple iterative processes, continuously decompose a problem into sub-problems, solve them one by one, and finally answer the original question.
We will provide you with a question and the previous iteration history (each iteration includes the thinking process, sub-question, the model's answers to the sub-questions based on its own knowledge or the relevant documents retrieved from the sub-question).  Please continue reasoning along the previous iteration history to solve the original problem. \\
\\
After reasoning, if you find you lack some knowledge, you can call a search engine by <search> query </search>, and it will return the top searched results between <information> and </information>.   If you find that no further external knowledge is needed, please still provide your reasoning process, and then you can directly provide the answers within <answer> and </answer>.  For example, <answer> Beijing </answer>. \\
\\
**Ensure both queries and answers are concise, using nouns or short phrases whenever possible.** \\
\\

Matters Need Attention:\\
**1. Please note that the previous iteration history is true and does not need to be verified again, so you can definitely take this information as established facts and, based on this, continue to reason downward.**\\
**2. Please note that during the reasoning process, do not end it casually. Only when you think it necessary to call the search engine (the output ends with </search>) or output the final answer (the output ends with </answer>) can the reasoning process be temporarily interrupted or completely ended.**\\

\\
\#\#\# Question \\
\{question\} \\
\\

\#\#\# Previous Iteration \\
\{iter\_history\} \\
\\

\#\#\# Your Output \\
\\
\bottomrule
\end{tabular}
\caption{The prompt for rollout.}
\label{tab_A.1-5}
\end{table*}

\end{document}